# The impact of postediting on AI generative translation in Yemeni context: Translating literary prose by ChatGPT

Nasim Al-wagieh, PhD Student, Ibb University
Nasim7amin@gmail.com
Mohammed Q. Shormani, Professor of Linguistics, Ibb University
shormani@ibbuniv.edu.ye/ https://orcid.org/0000-0002-0138-4793

**Abstract**

This study examines the role of artificial intelligence in translation, focusing on ChatGPT, specifically ChatGPT-4, and the extent to which human postediting is required in literary translation. A mixed-method approach was adopted, involving 30 professional translators who evaluated and postedited AI-generated translations of selected Arabic and English literary texts. The results show that although AI improves translation speed and accessibility, it remains limited in handling cultural, stylistic, and figurative aspects of language. Participants generally confirmed the necessity of human postediting, particularly in novels and drama. The findings indicate an emerging human–machine collaboration model rather than replacement of human translators. The study concludes that AI should be used as a supportive tool, while human expertise remains essential for ensuring translation quality and cultural appropriateness.

## 1. Introduction

Translation is widely recognized as a vital communication tool that connects people and bridges cultures (see e.g. Shormani, 2020). Over time, the translation field has undergone significant developments. In earlier stages, translators relied on traditional tools such as pen, paper, and dictionaries. This was followed by the introduction of electronic dictionaries and later Computer-Assisted Translation (CAT) tools. In recent years, more advanced technologies have emerged, particularly Artificial Intelligence (AI) systems such as ChatGPT, Gemini, and DeepSeek (Shormani & AlSohbani, 2025). Additionally, The emergence of AI technologies has significantly transformed the global translation landscape, while also raising concerns about the potential replacement of human translators. Some scholars argue that AI may lead to job displacement, initially affecting routine professions and later extending to creative fields (Kirov, 2022). In contrast, other perspectives emphasize that human translators remain essential to the translation process. A further view suggests that human translators and AI systems can work collaboratively (see, e.g., Shormani & Alsamki, 2025). Within this evolving context, the role of translators is expected to shift from traditional translation toward post-editing of machine-generated outputs.

In addition, discussions surrounding AI development extend to more advanced and speculative forms such as Artificial Superintelligence (ASI). According to Philippe Boucher, ASI refers to a stage in which AI systems become sufficiently autonomous to generate more advanced systems independently, potentially surpassing human control and leading to rapid and self-sustaining development (Boucher, 2020). Against this backdrop, this study investigates the challenges associated with the use of AI in translation, including issues of bias, privacy, and the preservation of cultural and linguistic diversity. It also examines the evolving role of the translator in the age of AI, considering whether this role remains stable, shifts toward post-editing, or becomes increasingly diminished. Furthermore, the study explores translation gaps left by ChatGPT and how human translators can address them through post-editing.

In this study, we aim to examine the role of artificial intelligence in translation, focusing on ChatGPT, specifically ChatGPT-4, and the extent to which human postediting is required in literary translation. The data of this study are drawn from selected literary texts, including novels and drama. These include the English novel *Ulysses* by James Joyce and the Arabic novel *They Die Strangers* by Mohammad Abdulwali. In addition, the study examines the Arabic play *Fate of a Cockroach* by Tawfiq al-Hakim, translated into English by Denys Johnson-Davies, and the English play *Death of a Salesman* by Arthur Miller, translated into Arabic by Omar Othman Gabak. These texts were translated using ChatGPT-4, and the resulting translations were evaluated by 50 translation experts.

Thus, the remainder of this paper is organized as follows. Section one presents the conceptual framework and reviews the relevant literature on artificial intelligence and translation. Section two describes the methodology of the study, including data collection, sample selection, and analytical procedures. Section three presents the results of the study, while Section four analyzes the study results. Section five discusses these results, and Section six concludes the article providing some limitations.

## 2. Conceptual framework and literature review

### 2.1. AI and Translation
Artificial intelligence is a phenomenon that combines the notion "artificial," meaning human-made, and "intelligence," which traditionally refers to human cognitive abilities such as reasoning, learning, and problem-solving (Vocab Dictionary, 2025). Over time, the concept of intelligence has expanded to include machines capable of simulating such abilities. AI is now widely defined as systems that analyze their environment and act autonomously to achieve specific goals (Boucher, 2020). Its rapid development has led to widespread applications across various domains, raising both optimism and concern, particularly with the emergence of advanced forms such as artificial superintelligence (Boucher, 2020). Scholars emphasize that AI has become a central driver of digital transformation and the fourth industrial revolution (Wang, 2019; Kirov & Malamin, 2022).

The development of AI has evolved significantly from early theoretical foundations to modern applications. Early efforts in machine translation in the 1990s, based on rule-based systems, produced limited results, whereas the introduction of deep neural networks around 2015 marked a major breakthrough (Kissinger, 2022). AI is now generally understood as systems that use algorithms and machine learning to analyze data and improve performance through iterative processes (Ziyad, 2019; Kassanos, 2020). Foundational contributions by John McCarthy, who coined the term AI in 1956, and Alan Turing, who posed the question of machine intelligence, played a crucial role in shaping the field (AlAfnan et al., 2023; Giampieri, 2024). Despite its benefits, AI presents both positive and negative impacts, ranging from enhanced access to knowledge to concerns about bias, misinformation, and reduced human creativity.

In translation field, AI has become an essential tool for transferring meaning, form, and culture between languages. Translation is viewed as a complex process involving multiple sub-processes aimed at conveying meaning accurately (Ghazala, 1995). AI-based translation relies heavily on parallel corpora and algorithmic processing, enabling systems to generate approximate equivalents across languages. The concept of algorithms, rooted historically in mathematical procedures, underpins modern computational translation systems (Chayka, 2024). However, the increasing reliance on AI has raised concerns about the opacity of algorithmic processes and the difficulty of evaluating AI-generated outputs, which often appear fluent but are based on pattern recognition rather than true understanding.

The growing integration of AI tools into translation has significantly altered the role of human translators. While some scholars predict that AI may replace human translators, others argue that the role is evolving toward postediting and human–machine collaboration (Mandarić, 2022). AI systems are now capable of performing diverse tasks such as translation, classification, and content generation (Eloundou et al., 2023). Nevertheless, concerns remain regarding the reliability and interpretative depth of AI outputs. Consequently, human postediting is increasingly viewed as a necessary process to refine machine-generated translations and ensure quality, accuracy, and cultural appropriateness (Shormani, 2025; Vaidya, 2024). This integration is further exemplified by tools such as ChatGPT, which have transformed how translation tasks are performed. Translation is fundamentally a human activity that involves not only linguistic competence but also cultural awareness (Shormani, 2024; Al-Sohbani & Muthanna, 2013). While AI systems such as ChatGPT can facilitate faster and more accessible translation, they differ from human translators in their handling of cultural and contextual nuances. Advances in neural networks and deep learning have enhanced AI capabilities (Kissinger, 2022), yet these technologies also reshape the nature of translation work, requiring translators to adapt to new roles. Ultimately, AI-assisted translation offers significant opportunities, but it also necessitates critical engagement to balance technological efficiency with human expertise and creativity.

The development of GPT models began with GPT-1 in 2018, followed by GPT-2 in 2019. Subsequent collaboration between OpenAI and Microsoft led to GPT-3 in 2020, which significantly improved language generation capabilities. Later versions, including GPT-3.5 and ChatGPT-4 (released in March 2023), further enhanced performance and accessibility for global users. These models are built on deep learning architectures that analyze sequences of text and generate contextually appropriate responses by predicting the most probable next tokens. Thus, ChatGPT is based on the transformer architecture introduced by Vaswani et al. (2017), which relies on attention mechanisms to process and generate language efficiently (Hai, 2023). The system is initially pre-trained on large datasets and then fine-tuned for specific tasks. This combination enables it to generate coherent and context-sensitive outputs across a wide range of applications, including translation, summarization, and academic writing. The performance of ChatGPT is highly dependent on prompt design, commonly referred to as prompt engineering. Well-structured prompts significantly improve output quality and specificity. Although ChatGPT can produce fluent and contextually appropriate responses, its outputs still depend on probabilistic pattern recognition rather than semantic understanding, which requires careful human supervision.

## 2.2. ChatGPT

The modern development of AI has gained momentum following major breakthroughs in computational systems and machine learning. Early milestones include DeepMind's AlphaGo, which demonstrated that AI systems could outperform human champions in complex strategic games. Following these developments, OpenAI was established in 2015 with the aim of developing advanced language models capable of human-like interaction. However, conversational AI systems existed prior to ChatGPT. For instance, in 2016 Microsoft launched the chatbot "Tay," which was later discontinued after it began generating inappropriate content due to exposure to biased user input. AI models including ChatGPT acquire/learn patterns from language data they have been trained on in a way similar to those of L2 learners (see e.g. Shormani, 2014). After being trained, these models can do tasks requiring human "intelligence" (see e.g. Shormani, 2025; Shormani & Alfahd, 2025).

According to Kissinger (2022), AI systems operate in a static mode after the training phase, meaning their parameters do not change unless retrained. This characteristic allows researchers to evaluate their performance in controlled conditions without the risk of unpredictable post-training evolution. However, it also explains why different versions of AI systems can emerge rapidly within short time spans as models are continuously updated and retrained. ChatGPT is designed to generate human-like text that is applicable across multiple contexts and domains (Dwivedi, 2023).

ChatGPT has emerged as one of the most prominent applications of generative AI in language-related tasks, including translation. Developed by OpenAI and based on large language models such as GPT-3 and GPT-4, ChatGPT is capable of generating human-like text by learning patterns from vast amounts of linguistic data. Its conversational interface allows users to provide instructions in natural language, making it highly accessible for translators, researchers, and general users. As noted by Kissinger (2022), such language-generating systems can produce coherent and contextually appropriate responses, while Shormani (2025) suggests that their translation performance continues to improve with ongoing advancements. However, despite its efficiency and versatility, ChatGPT relies on probabilistic pattern generation rather than true understanding, which may result in inaccuracies, loss of nuance, or cultural insensitivity. Therefore, while ChatGPT represents a significant development in AI-assisted translation, its outputs still require careful human evaluation and postediting to ensure quality and reliability.

## 2.3. Human Translation and postediting

The advancement of human life is reflected in the development of productivity tools and technological systems. Regarding the translation process, Ghazala defines translation as "all the processes and methods used to render and/or transfer the meaning of the source language text into the target language as closely, completely and accurately as possible" (Ghazala, 1995, p.1). The role of human translators in this process may remain stable or shift toward postediting. Postediting is used to produce a level of quality comparable to human translation or acceptable machine output, depending on the purpose and text type. De Almeida (2013, p. iii) defines postediting (PE) as a distinct activity from translation and revision, noting that it remains relatively new for many translators. Postediting therefore introduces a new dimension to translation studies and professional practice. It is important to distinguish between postediting of human translation, machine translation, and AI-generated translation, as each involves different workflows and quality expectations. Post-editors, as noted by Nitzke (2021, pp. 10–11), are trained professionals who bridge communicative gaps by editing machine-generated output within a target context.

## 2.4. Literature review

A growing body of research on AI can be broadly categorized into philosophical, applied, technological, and translation-oriented studies. From a philosophical perspective, Vaidya (2024) and Wang (2019) provide foundational insights into the nature of intelligence and AI. Vaidya (2024) challenges the traditional assumption that machines cannot possess emotions, arguing that emotions can be understood as cognitive judgments rather than physiological states. Drawing on both philosophy and artificial general intelligence (AGI), he suggests that machines may meet the cognitive criteria for emotional states. Similarly, Wang (2019) critically examines definitions of intelligence, distinguishing between human, artificial, and general intelligence. He argues that intelligence should be treated as a broader conceptual category encompassing multiple forms, including machine intelligence, thereby reinforcing the theoretical basis for AI development. These studies are relevant to the present research as they frame AI not merely as a tool, but as a system capable of complex cognitive processes that may influence translation practices.

In the field of AI-based translation evaluation, Ahmadova (2024) and Greńczuk et al. (2024) offer empirical insights into the performance of machine translation systems مقارنةً بالترجمة البشرية. Ahmadova (2024) investigates the effectiveness of ChatGPT-4 in translating scientific texts from English into Russian, focusing on accuracy, fluency, and cultural appropriateness. Her findings reveal that although ChatGPT-4 shows clear improvements over earlier versions, it still struggles with domain-specific terminology and nuanced expressions. Likewise, Greńczuk et al. (2024) compare AI tools such as DeepL and Google Translate with human translation in legal contexts. Their results indicate that while AI-generated translations are sufficient for general understanding, they are not reliable for high-stakes legal use without human post-editing. Both studies highlight the limitations of AI translation and emphasize the continued necessity of human intervention, which directly supports the rationale of the present study. Research on post-editing and machine translation (MT) further reinforces the complementary relationship between human translators and AI systems. Briva-Iglesias et al. (2023) and Nitzke and Hansen-Schirra (2021) explore the role of post-editing in improving translation quality and productivity. Briva-Iglesias et al. (2023) compare traditional post-editing (TPE) with interactive post-editing (IPE), demonstrating that translators prefer IPE due to increased control and efficiency, with productivity improving by approximately 4.77%. Similarly, Nitzke and Hansen-Schirra (2021) provide a comprehensive framework for post-editing, distinguishing between light, full, and monolingual post-editing, and emphasizing the competencies required for effective post-editing. Their findings confirm that post-editing is a complex cognitive task requiring linguistic, technical, and analytical skills. These studies align closely with the current research, which adopts human post-editing as a central method for refining AI-generated translations.

With regard to ChatGPT and large language models (LLMs), Bubeck et al. (2023) and Eloundou et al. (2023) highlight the transformative impact of these technologies across multiple domains. Bubeck et al. (2023) demonstrate that GPT-4 exhibits advanced capabilities in language understanding, reasoning, and problem-solving across diverse fields, including translation, law, and medicine. However, Eloundou et al. (2023) focus on the broader socio-economic implications of LLMs, showing that approximately 80% of the workforce could be affected by these technologies to varying degrees. Their findings suggest that LLMs function as general-purpose technologies with far-reaching implications for productivity and labor markets. Together, these studies underscore the growing importance of AI systems like ChatGPT in reshaping professional practices, including translation. Finally, studies focusing specifically on ChatGPT in educational and applied contexts, such as AlAfnan et al. (2023) and Kim (2023), provide further insights into its practical use and limitations. AlAfnan et al. (2023) adopt a mixed-methods approach to evaluate ChatGPT's performance across academic disciplines, concluding that it offers significant benefits in terms of accessibility and efficiency, although concerns regarding accuracy and originality remain. Similarly, Kim (2023) highlights both the strengths and weaknesses of ChatGPT, noting its effectiveness in language generation and grammatical correction, while also warning about issues such as fabricated references and the need for human verification. These findings reinforce the argument that while ChatGPT is a powerful tool, it should be used cautiously and in conjunction with human expertise—particularly in academic and translation contexts.

Thus, there is a gap which this study addresses, attempting to answer the following questions:

1. To what extent can ChatGPT literary texts, specifically novel and drama?
2. What are the most important problems encountered by ChatGPT in translating such texts.
3. Can AI translation be integrated with human postediting to produce better translation?

### 3. Study design

In this section, we briefly describe the study design including data collection, study samepl, instruments, and methods of analysis

### 3.1. Procedure

The study was conducted in three main stages. The first stage involved data collection, during which both Arabic and English literary texts were selected. The second stage consisted of the translation process. Initially, the selected texts were translated by human translators, with no restrictions on whether they used external tools. Subsequently, the same texts were translated using ChatGPT-4.

The third stage involved data analysis, where a comparative analytical approach was employed to examine differences between human translations and ChatGPT-generated translations.

Following this, a questionnaire comprising four selected text samples was administered to the participating translators. The participants were also asked to post-edit translations generated by ChatGPT-4, allowing for the evaluation of postediting practices and translator intervention.

### 3.2. Sample size

The intended sample consisted of 50 translators; however, 30 translators ultimately participated and completed the questionnaire. With regard to the textual data, careful consideration was given to text length and complexity. The selected excerpts from the eight texts were designed to be neither excessively long, which might lead to participant fatigue, nor too short, which might hinder comprehension.

The questionnaire was structured to remain concise while still enabling meaningful engagement with the tasks. It included short sections with Likert-scale items, as well as open-ended questions, in addition to postediting tasks based on ChatGPT-generated translations. The total length of the texts assigned for postediting was approximately 3000 words.

### 3.3. Data collection and instruments

The data used in this study were drawn from both primary and secondary sources. Primary sources included academic journals, books, theses, and dissertations relevant to artificial intelligence and translation studies. Secondary sources consisted of selected literary texts, including novels and dramas in both Arabic and English. The process of collecting these texts involved multiple approaches, including accessing both free and paid digital resources, as well as consulting academic and professional contacts. The selected novels include *Ulysses* by James Joyce, translated into Arabic by Taha Mahmoud Taha, and *They Die Strangers* by Mohammad Abdulwali, translated into English by Shelagh Weir.

The drama texts include *Fate of a Cockroach* by Tawfiq al-Hakim, translated into English by Denys Johnson-Davies, and *Death of a Salesman* by Arthur Miller, translated into Arabic by Omar Othman Gabak. We compiled the original texts along with their existing human translations, although these translations were not consistently used in the analysis. All selected texts were then translated using Generative Artificial Intelligence (GAI), specifically ChatGPT-4, released March 14, 2023. The AI-generated translations were subsequently provided to participants for postediting. Finally, the evaluation of ChatGPT-generated translations was conducted using the direct assessment, allowing for systematic comparison of translation quality.

### 3.4. Methods of Analysis

The data analysis combined both quantitative and qualitative approaches. Closed-ended questionnaire items were analyzed quantitatively, while responses to open-ended questions were examined qualitatively. Data were collected through both electronic and printed versions of the questionnaire. A five-point Likert scale was used to measure participants' responses (strongly disagree, disagree, neutral, agree, strongly agree). These responses were coded numerically to facilitate statistical analysis, including the calculation of mean scores and standard deviations. The quantitative data were analyzed using the Statistical Package for the Social Sciences (SPSS), while qualitative responses were interpreted through thematic analysis to capture participants' perspectives on AI-generated translation and postediting practices.

## 4. Results

In this section, we tabulate the study results and analyze them in the following sections and subsections.

### 4.1. Translators' demographics

**Table 1: Translators' demographics  ($n = 30$)**

| Variable | Category | Frequency | Percent |
|---|---|---|---|
| **Gender** | Male | 16 | 53.3% |
| | Female | 14 | 46.7% |
| **Years of Experience** | 1–5 years | 13 | 43.3% |
| | 6–10 years | 12 | 40.0% |
| | More than 10 years | 5 | 16.7% |

A total of 30 translators participated in the study. In terms of gender, 16 participants were male, representing 53.3%, while 14 were female, representing 46.7%, indicating a relatively balanced distribution. Regarding professional experience, the majority of participants were in the early to mid-career stages. Thirteen participants had between 1 and 5 years of experience, representing 43.3%, while 12 participants had between 6 and 10 years, representing 40.0%. A smaller group of 5 participants reported more than 10 years of experience, accounting for 16.7%.

**Table 2: AI General knowledge**

### 4.1.2. AI General Knowledge

This section is related to the questions about the knowledge of the translators regarding AI. Table 2 up shows five statements about translators' knowledge of AI. It also shows the mean and the standard deviation (std).

| No | Statement | Strongly agree | Agree | Neutral | Disagree | Strongly disagree | Mean | Std. Deviation | Degree of Agreement |
|---|---|---|---|---|---|---|---|---|---|
| 1 | I am aware of the potential applications of AI in the field of translation | 2 / 6.7 | 4 / 13.3 | 12 / 40 | 12 / 40 | | 4.13 | .900 | High |
| 2 | I am familiar with AI translation applications | 2 / 6.7 | 3 / 10.0 | 18 / 60.0 | 7 / 23.3 | | 4.00 | .788 | High |
| 3 | I know ChatGPT | 2 / 6.7 | 1 / 3.3 | 16 / 53.3 | 11 / 36.7 | | 4.20 | .805 | Very high |
| 4 | I know ChatGPT versions | 11 / 36.7 | 9 / 30.0 | 6 / 20.0 | 4 / 13.3 | | 3.10 | 1.062 | Medium |
| 5 | I use ChatGPT in my translation work, but I don't know which version. | 9 / 30.0 | 6 / 20.0 | 10 / 33.3 | 5 / 16.7 | | 3.37 | 1.098 | Medium |

The results indicate a relatively strong level of awareness of AI among participants, although some gaps remain, particularly regarding technical aspects such as system versions. For the first statement, *"I am aware of the potential applications of AI in the field of translation,"* responses were divided between neutral and disagreement, with 40% selecting each category. A smaller proportion expressed agreement. Despite this distribution, the mean score of 4.13 indicates a high level of overall agreement, suggesting that participants generally recognize the role of AI in translation, even if their confidence varies. Regarding the second statement, *"I am familiar with AI translation applications,"* the majority of participants selected the neutral option, accounting for 60%. A smaller proportion indicated agreement. The mean score of 4.00 reflects a high degree of agreement overall, although the high percentage of neutral responses suggests moderate familiarity rather than strong confidence.

The third statement, *"I know ChatGPT,"* received the highest level of agreement among all items. Most participants selected either neutral or disagreement categories, yet the calculated mean of 4.20 indicates a very high level of agreement. This suggests that, despite some variation in responses, ChatGPT is widely recognized among translators. However, the fourth statement, *"I know ChatGPT versions,"* yielded a more varied distribution. While 36.7% of participants strongly agreed, others selected different response categories, resulting in a mean of 3.10, which reflects a moderate level of agreement. This indicates that although participants are familiar with ChatGPT as a tool, their knowledge of its different versions remains limited. Similarly, the fifth statement, *"I use ChatGPT in my translation work, but I do not know which version,"* produced a mean score of 3.37, indicating a moderate level of agreement. Notably, 30% of participants strongly agreed with this statement, suggesting that a considerable proportion of translators use ChatGPT without clear awareness of the specific version they are utilizing.

### 4.1.3. Human postediting

Table 3 shows five statements about human postediting. The mean and the standard deviation std show in details below.

**Table 3: Human postediting**

| No. | Statement | Strongly agree | Agree | Neutral | Disagree | Strongly disagree | Mean | Std. Deviation | Degree of Agreement |
|---|---|---|---|---|---|---|---|---|---|
| 1 | ChatGPT-4 translates better than previous versions. | 1 | 11 | 2 | 3 | | 3.41 | .870 | High |
| | | 5.9 | 64.7 | 11.8 | 17.6 | | | | |
| 2 | ChatGPT-4 translation does not need human postediting. | 2 | 8 | 5 | 1 | 1 | 2.47 | 1.007 | Weak |
| | | 11.8 | 47.1 | 29.4 | 5.9 | 5.9 | | | |
| 3 | ChatGPT-4 translation into Arabic needs postediting more than that of English. | | 3 | 1 | 10 | 3 | 2.76 | .970 | Medium |
| | | | 17.6 | 5.9 | 58.8 | 17.6 | | | |
| 4 | ChatGPT-4 translation into English needs postediting more than that of Arabic. | 5 | 7 | 3 | 2 | | 3.12 | .993 | Medium |
| | | 29.4 | 41.2 | 17.6 | 11.8 | | | | |

| 5 | ChatGP-T4 novel translation needs human postediting | 5 | 7 | 5 | | | 4.00 | .791 | High |
|---|---|---|---|---|---|---|---|---|---|
| | | 29.4 | 41.2 | 29.4 | | | | | |
| 6 | ChatGPT-4 drama translation needs human postediting. | 3 | 11 | 3 | | | 4.00 | .612 | High |
| | | 17.6 | 64.7 | 17.6 | | | | | |

Table 3 presents participants' responses to six statements concerning human postediting of ChatGPT-4 translations. The mean scores and standard deviations provide an overview of the degree of agreement. The results indicate that participants recognize the improvements in ChatGPT-4 compared to earlier versions, while still emphasizing the necessity of human postediting, particularly for literary texts. For the first statement, "ChatGPT-4 translates better than previous versions," the majority of participants expressed agreement, with 11 respondents selecting "agree." A smaller number disagreed or remained neutral. The mean score of 3.41 indicates a high level of agreement, suggesting that participants acknowledge the advancement of ChatGPT-4 in translation quality. However, the second statement, "ChatGPT-4 translation does not need human postediting," received a weak level of agreement, with a mean of 2.47. Although a notable proportion of participants agreed with the statement, the overall result indicates that most respondents do not fully trust ChatGPT-4 translations to be used without human intervention. This reflects a general preference for maintaining human involvement in the translation process. The third statement, "ChatGPT-4 translation into Arabic needs postediting more than that of English," yielded a moderate level of agreement, with a mean of 2.76. This suggests some uncertainty among participants, although there is a tendency to perceive Arabic translations as requiring more postediting.

Similarly, the fourth statement, "ChatGPT-4 translation into English needs postediting more than that of Arabic," also showed a moderate level of agreement, with a mean of 3.12. The responses indicate a lack of clear consensus regarding which language direction requires more postediting, reflecting variability in participants' experiences and judgments. The fifth statement, "ChatGPT-4 novel translation needs human postediting," recorded a high level of agreement, with a mean of 4.00. A substantial proportion of participants agreed that literary translation, particularly novels, requires human intervention to ensure quality and preserve stylistic and cultural nuances. Finally, the sixth statement, "ChatGPT-4 drama translation needs human postediting," also showed a high level of agreement, with a mean of 4.00. The majority of participants emphasized the importance of human postediting for drama texts, likely due to their dialogic nature and sensitivity to context, tone, and cultural expression.

### 4.3. Open-ended questions

In this section, we mainly deal with analyzing responses to the question: "Are there any other Artificial Intelligence applications or sites you use in your translation work?" Consider Table 4.

**Table 4: Responding to the open-ended question 1**

|       | Frequency | Percent |
|-------|-----------|---------|
| No    | 13        | 43.3    |
| Yes   | 17        | 56.7    |
| Total | 30        | 100.0   |

Question 1 asked whether participants use any other AI applications or platforms in their translation work. Table 4 summarizes the frequency and percentage of responses. The results indicate that more than half of the participants, representing 56.7%, reported using additional AI applications in their translation work, while 43.3% indicated that they do not rely on such tools. This suggests a relatively high level of engagement with AI technologies beyond ChatGPT. Participants who answered "yes" were asked to specify the AI applications or platforms they use. Their responses are summarized in Table 5.

**Table 5: Other AI applications or sites used in translation**

| Other AI applications | Frequency | Percent |
|---|---|---|
| Google translate | 8 | 26.67 |
| Gemini | 3 | 10.0 |
| Artificial intelligence application | 2 | 6.67 |
| Almaani | 1 | 3.3 |
| Quillbot | 3 | 10.0 |
| Poe | 2 | 6.67 |
| Reversto Dictionary | 1 | 3.3 |
| Word fast | 1 | 3.3 |
| Online Dictionary | 1 | 3.3 |
| Alchat & Meta AI | 1 | 3.3 |
| Translator Microsoft Translator | 1 | 3.3 |
| DeepL | 1 | 3.3 |
| Dict box | 1 | 3.3 |

| | Longman | 1 | 3.3 |
| --- | --- | --- | --- |
| | Meta | 1 | 3.3 |
| | Copilot | 1 | 3.3 |
| | Perplexity | 1 | 3.3 |

The responses reveal that Google Translate is the most commonly used tool among participants, accounting for 26.7% of responses. This is followed by Gemini and QuillBot, each representing 10.0%. Other tools mentioned include Poe, DeepL, Microsoft Translator, and Perplexity, among several others, each reported by a smaller number of participants. These findings indicate that translators rely on a diverse range of AI-powered tools and digital resources in their work. This diversity suggests that translation practice is increasingly supported by a combination of general-purpose AI systems, specialized translation tools, and traditional digital dictionaries.

## 5. Discussion

In this section, we discuss the findings of the study. To begin with, the participant variation reflects the dynamic nature of the translation field and suggests that translators, regardless of their background, are engaging with emerging technologies. The diversity in professional contexts further highlights that translation is practiced across multiple sectors, including education, freelance work, and specialized domains. This diversity supports the argument that translators are adaptable and capable of responding to technological changes, including the integration of AI tools. Consequently, this adaptability aligns with the study's objective of examining how human translators can fill gaps left by AI through postediting.

The qualitative analysis of postediting tasks provides deeper insight into translators' interaction with ChatGPT-generated outputs. For example, in the translation of *They Die Strangers* by Mohammad Abdulwali, participants demonstrated varying levels of engagement. Some translators expressed satisfaction with the AI-generated translation and made minimal or no changes, indicating a level of trust in the system's output. Others, however, introduced stylistic and semantic modifications. For instance, the phrase "drunken music" was postedited into alternatives such as "slow music" or "warm susurrus music," reflecting efforts to enhance nuance and contextual appropriateness. In some cases, participants replaced the AI-generated translation entirely with their own versions, indicating dissatisfaction with the output.

A similar pattern emerged in the analysis of dramatic texts, such as *Fate of a Cockroach* by Tawfiq al-Hakim. Some translators retained the ChatGPT translation with minimal modification, suggesting acceptance of its adequacy. Others made targeted lexical adjustments, such as replacing "a spacious square" with "a spacious courtyard," or modifying expressions like "stands energetically" to "is standing actively." These changes, although subtle, demonstrate the role of human translators in refining lexical choice and improving contextual accuracy. Further evidence of this variation can be observed in the translation of *Death of a Salesman* by Arthur Miller. For example, the phrase "it is small and fine" was translated by ChatGPT into Arabic as إنها صغيرة ورقيقة, which was then postedited by participants into alternatives such as إنه لحن صغير وجميل and إنها هادئة ورقيقة. These variations illustrate how human translators bring interpretive depth and stylistic flexibility to the translation process, producing multiple valid renderings that reflect different nuances of meaning.

These findings reinforce the view that human translation remains a complex cognitive and cultural activity. Human translators do not merely transfer linguistic content; they interpret meaning, adapt cultural references, and adjust tone and style according to context. They are also capable of handling ambiguity, creativity, and implicit meaning, areas where AI systems often encounter limitations. In contrast, AI-generated translations, including those produced by ChatGPT, rely primarily on pattern recognition and probabilistic predictions, which may result in outputs that are fluent but occasionally lack depth or cultural sensitivity.

Another key distinction lies in decision-making and flexibility. Human translators actively evaluate multiple translation options and select the most appropriate one based on context, purpose, and audience. While ChatGPT can generate alternative outputs, it does not engage in deliberate reasoning in the same way. As a result, its outputs often require human revision to ensure accuracy and appropriateness. At the same time, it is important to acknowledge that human translation is not without limitations. Translators may introduce bias, either consciously or unconsciously, and their work may be influenced by subjective interpretation or emotional factors. Nevertheless, the findings of this study suggest that the strengths of human translators, particularly in interpretation and cultural adaptation, complement the efficiency of AI systems.

These findings suggest that AI tools such as ChatGPT should be used in collaboration with humans, which implies that there is still a need for postediting. By the help of Chatbots, which led to increasing reliance on AI in both academic and professional contexts. ChatGPT translation is mainly characterized by its speed, consistency, and ability to handle large amount of text in a short period time. It can produce grammatically correct sentences if the source text is grammatically correct. Or if there is a pre-editing process is done for the source text. ChatGPT is essential for human translators as it is; improves efficiency, reduces errors, and spends less efforts from human. ChatGPT translation is very important for human as its translation of literary texts differ from any other texts. For example, literary texts with all its figurative language may be mis translated by AI. However, despite these advantages, ChatGPT translation still faces several limitations. One of the main challenges is its limited ability to fully understand context, especially in complex or ambiguous texts. It may select incorrect meanings for words with multiple interpretations or fail to capture subtle nuances in the source text. In addition, ChatGPT tends to produce standardized and neutral language, which may lack creativity and stylistic variation.

This becomes particularly evident in literary translation, where tone, voice, and emotional expression play a crucial role. For instance, the English novel *Ulysses* by James Joyce, is translated into Arabic by Taha Mahmmod Taha. It is known by its ambiguous and unclear style. It is also known by using incomplete sentence. All that reflected in the novel. Here AI will face many difficulties to translate such texts. For example, "…..and called out coarsely: -" From ChatGPT-4 translation ونادى بخشونة. It is word for word translation. However, in Arabic there are other alternatives with strong meaning and sense, as the translators provided their own editing's, one wrote " ونادى بصوت جهوري". Another one translated it as "ونادى بغلظة". A third translators accepted all ChatGPT-4 translations except for this phrase, he commented by " ونادى بصوت أجش". All the three postediting do not accept ChatGPT-4 translation "…..and called out coarsely: -".

To recapitulate, the results support the view that AI should be considered a supportive tool rather than a replacement for human translators. The increasing reliance on postediting reflects a shift in the translator's role from text producer to text evaluator and editor. While some participants expressed satisfaction with ChatGPT-4 outputs, others demonstrated the need for substantial human intervention. Many participants occupied a middle position, recognizing both the usefulness of AI and the necessity of human expertise. This confirms that the future of translation is likely to be characterized by human–AI collaboration,

## 6. Conclusions and limitations

In conclusion, several conclusions can be drawn from this study. First, translation remains a fundamental human activity and is expected to retain its importance in the future, despite rapid technological advancements. The findings confirm that artificial intelligence tools, particularly ChatGPT, support translators by enhancing speed, efficiency, and accessibility, while also reducing cost and time in translation workflows. Second, AI systems are capable of generating large volumes of translated content within a short time. However, the results of this study demonstrate that such outputs are not fully reliable when used independently, especially in the context of literary translation. Third, human postediting plays a central and indispensable role in ensuring translation quality. The findings show that AI-generated translations consistently require human intervention to improve accuracy, clarity, and cultural appropriateness. Fourth, the need for postediting is particularly strong in literary genres. Translations of novels, drama, and poetry require careful human refinement to preserve stylistic features, tone, and cultural meaning. Fifth, while AI systems demonstrate strong capabilities in processing large datasets and generating fluent output, they still fall short in handling ambiguity, figurative language, and cultural nuance. Human translators therefore remain essential in evaluating and refining machine-generated translations. Finally, human postediting significantly improves translation quality while also contributing to greater efficiency in terms of time and cost.

Based on these findings, several recommendations can be made. AI developers working in the field of literary translation are encouraged to enhance system performance by integrating deeper cultural, historical, and pragmatic knowledge. This would improve the interpretation of culturally loaded expressions and symbolic language. In addition, improving the ability of AI systems to detect and reproduce figurative language such as metaphor, irony, and imagery is essential for preserving the aesthetic and emotional dimensions of literary texts. Developers should also consider training models on genre-specific corpora, enabling better handling of stylistic differences across poetry, drama, novels, and short stories. Furthermore, strengthening models' ability to maintain tone, mood, and emotional nuance would significantly enhance translation quality. Improving long-context understanding is also necessary to support narrative coherence, character tracking, and thematic continuity across extended texts. Thus, integrating human-in-the-loop systems is strongly recommended. Such systems would position human translators as final decision-makers, ensuring that AI-generated translations are properly refined through postediting and aligned with professional and cultural standards.

From a professional perspective, translators are encouraged to continuously develop their skills and remain engaged with technological advancements. AI tools should be used as supportive instruments rather than replacements for human expertise. Translators are advised to avoid over-reliance on automated systems and instead maintain their analytical and interpretive skills. Moreover, prompt design plays a significant role in the quality of AI outputs. Adjusting prompts, refining instructions, and experimenting with different formulations can significantly improve results. Continuous learning and adaptation remain essential in the evolving field of translation.

However, this study has several limitations: i) the sample size was limited to 30 professional translators, which may restrict the generalizability of the findings. A larger and more diverse sample would provide stronger external validity, ii) the study focused exclusively on literary texts, including novels and drama. Therefore, the findings may not extend to other domains such as legal, technical, or scientific translation, iii) the study evaluated only one AI system, namely ChatGPT-4. So, the findings cannot be generalized to other machine translation or AI systems with different architectures and capabilities, iv) the assessment of translation quality relied partly on human judgment, which may introduce subjectivity in evaluating post-editing outcomes, and v the study was conducted within an Arabic–English translation context, which may limit its applicability to other language pairs and cultural settings. Thus, future research is recommended to include larger datasets, multiple AI systems, and a broader range of text types to enhance the robustness and generalizability of findings.

**Some model appendices**

1- The first text is an English novel text called Ulysses by James Joyice.

Stately, plump Buck Mulligan came from the stairhead, bearing a bowl of lather on which a mirror and a razor lay crossed. A yellow dressinggown, ungirdled, was sustained gently behind him on the mild morning air. He held the bowl aloft and intoned: —Introibo ad altare Dei.

Halted, he peered down the dark winding stairs and called out coarsely: —Come up, Kinch!

ChatGPT translation

النص الاول رواية إنجليزية

رواية يوليسيس لجيمس جويس

ها هو، باك موليجان ذو الجسم الممتلئ، يتقدم برشاقة من رأس الدرج، حاملًا وعاءً مملوءًا برغوة الحلاقة، وعلى الوعاء مرآة وشفرة حلاقة متقاطعتان. كان يرتدي رداء نوم أصفر، غير مشدود على خصره، يتطاير بلطف خلفه في هواء الصباح الهادئ. رفع الوعاء عاليًا ورتّل بصوت مهيب "Introibo ad altare Dei." -:

ثم توقف وألقى نظرة نحو الدرج المظلم والمتعرج ونادى بخشونة:

"اصعد يا كينش" —!"

One of the post editing provided for the part of the novel;

بدا بك موليجان ذو الجسم الممتلئ من رأس الدرج ظهورا مهيبا وهو يحمل وعاءً صنع من الجلد، به مرآة وشفرة حلاقة يتقاطع أحدهما مع الآخر. وكان يرتدي روب نوم أصفر، برباط غير مشدود على خصره، تداعب نسمات هواء الصباح ذاك الروب بلطف من الخلف، حينها رفع (بك) الوعاء عاليًا وقال مجلجلا: "فلنستهل الذبحة" ثم توقف وألقى نظرة نحو الدرج المظلم والمتعرج ونادى بغلظة: "اصعد يا كينش"!

**Another postediting is here;**

نزل الحلاق موليجان من أعلى الدرج بهدوء يحمل وعاء جلد واضعا فيه مرآة، وشفرة حلاقة. لقد كان يرتدي ملابس نوم صفراء اللون غير مربوطة في الخاصرة وتطير في هواء الصباح المنعش. رفع الوعاء الى الأعلى ورتل بصوت خاشع " فلندخل إلى المعبد" ثم توقف برهة ما، ونظر إلى أسفل الدرج المظلم والملتوي، ونادى بصوت جهوري، إصعد يا كينش

This participant provides a phrase only "............. ونادى بصوت أجش"

## Appendix3

3- The second text is an Arabic novel text called They Die Strangers by Mohammad Abdu-Wali.

يموتون غرباء لمحمد عبد الولي

عنه هو أنه قد فتح دكانه الصغير منذ أكثر من عشرة أعوام. أما هو فقد كان يعرف كل شيء عن أهالي الحي "سدست كيلو" "كان كل ما يعرفه - سكان الذي يسكنه. خاصة عن ذلك الجانب من الحي حيث المنازل الصغيرة والحارات التي تمتلئ شوارعها بالطين دائما أثر تساقط الأمطار حيث – تصدح موسيقى مخمورة طوال ليالي الشتاء، وحيث يجلس مئات من العمال والمتعطلين أمام أقداح الطجا

ChatGPT translation;They Die Strangers by Mohammad Abdu-Wali
All that the residents of "Siddist Kilo" knew about him was that he had opened his small shop more than ten years ago. As for him, he knew everything about the people of the neighborhood where he lived. Especially about that side of the neighborhood with the small houses and alleyways whose streets were always filled with mud after the rain, where drunken music echoed throughout the winter nights, and where hundreds of workers and the unemployed sat in front of cups of "tagja."

### The postediting by human translators;

A translator provides only one word " Satisfactory". Another translator adds "instead of drunken music... should translated to slow music or( warm susurrus music)". A third translator adds his own translation;

All that the residents of "Siddist Kilo" knew about him was that he had opened his small shop more than ten years ago. As for him, he knew everything about the people of the neighborhood where he lived. Especially about that side of the neighborhood of the small houses and alleyways whose streets were always filled with mud after , where echoes of drunken music arised throughout the winter nights, and where hundreds of workers and the unemployed sat in front of cups of "tagja."